\documentclass[times,twocolumn,final]{elsarticle}

\usepackage{prletters}
\usepackage{framed,multirow}

\usepackage{amssymb}
\usepackage{latexsym}

\usepackage{url}
\usepackage{xcolor}
\usepackage{color,hyperref}
\usepackage{bm}
\usepackage{amsmath}

\def\x{\bm{x}}

\def\W{\bm{W}}
\def\u{\bm{u}}
\def\s{\bm{s}}
\def\W{\bm{W}}
\def\x{\bm{x}}

\definecolor{newcolor}{rgb}{.8,.349,.1}

\journal{Pattern Recognition Letters}

\begin{document}

\thispagestyle{empty}

\ifpreprint
  \setcounter{page}{1}
\else
  \setcounter{page}{1}
\fi

\begin{frontmatter}

\title{COVID-CAPS: A Capsule Network-based Framework for Identification of COVID-19 cases from X-ray Images}

\author[1]{Parnian {Afshar}}
\author[2]{Shahin {Heidarian}}
\author[1]{Farnoosh {Naderkhani}}
\author[3]{Anastasia {Oikonomou}}
\author[4]{Konstantinos N.  {Plataniotis}}
\author[1]{Arash {Mohammadi}\corref{cor1}}
\cortext[cor1]{The paper is under consideration at Pattern Recognition Letters.}
\ead{arash.mohammadi@concordia.ca}

\address[1]{Concordia Institute for Information Systems Engineering, Concordia University, Montreal, QC, Canada}
\address[2]{Department of Electrical and Computer Engineering, Concordia University, Montreal, QC, Canada}
\address[3]{Department of Medical Imaging, Sunnybrook Health Sciences Centre, University of Toronto, Canada}
\address[4]{Department of Electrical and Computer Engineering, University of Toronto, Toronto, ON, Canada}

\begin{abstract}
Novel Coronavirus disease (COVID-19) has abruptly and undoubtedly changed the world as we know it at the end of the 2nd decade of the 21st century. COVID-19 is extremely contagious and quickly spreading globally making its early diagnosis of paramount importance. Early diagnosis of COVID-19 enables health care professionals and government authorities to break the chain of transition and flatten the epidemic curve. The common type of COVID-19 diagnosis test, however, requires specific equipment and has relatively low sensitivity. Computed tomography (CT) scans and X-ray images, on the other hand, reveal specific manifestations associated with this disease.  Overlap with other lung infections makes human-centered diagnosis of COVID-19 challenging. Consequently, there has been an urgent surge of interest to  develop Deep Neural Network (DNN)-based diagnosis solutions, mainly based on Convolutional Neural Networks (CNNs), to facilitate identification of positive COVID-19 cases. CNNs, however, are prone to lose spatial information between image instances and require large datasets. The paper presents an alternative modeling framework based on Capsule Networks, referred to as the COVID-CAPS, being capable of handling small datasets, which is of significant importance due to sudden and rapid emergence of COVID-19. Our results based on a dataset of X-ray images show that COVID-CAPS has advantage over previous CNN-based models. COVID-CAPS achieved an Accuracy of $95.7\%$, Sensitivity of $90\%$, Specificity of $95.8\%$, and Area Under the Curve (AUC) of $0.97$, while having far less number of trainable parameters in comparison to its counterparts. To potentially and further improve diagnosis capabilities of the COVID-CAPS, pre-training and transfer learning are utilized based on a new dataset constructed from an external dataset of X-ray images. This is in contrary to existing works where pre-training is performed based on natural images. Pre-training with a dataset of similar nature further improved accuracy to $98.3\%$ and specificity to $98.6\%$. 
\end{abstract}


\end{frontmatter}


\section{Introduction}
Novel Coronavirus disease (COVID-19), first emerged in Wuhan, China~\cite{Xu:2020}, has abruptly and significantly changed the world as we know it at  the end of the 2nd decade of the 21st century. COVID-19 seems to be extremely contagious and quickly spreading globally with common symptoms such as fever, cough, myalgia, or fatigue resulting in ever increasing number of human fatalities. Besides having a rapid human-to-human transition rate, COVID-19 is associated with high Intensive Care Unit (ICU) admissions resulting in an urgent quest for development of fast and accurate diagnosis solutions~\cite{Xu:2020}. Identifying positive COVID-19 cases in early stages helps with isolating the patients as quickly as possible~\cite{Wang:2020}, hence  breaking the chain of transition and flattening the epidemic curve.

Reverse Transcription Polymerase Chain Reaction (RT-PCR), which is currently the gold standard in COVID-19 diagnosis~\cite{Xu:2020}, involves detecting the viral RNA from sputum or nasopharyngeal swab. The RT-PCR test is, however, associated with relatively low sensitivity (true positive rate) and requires specific material and equipment, which are not easily accessible~\cite{Xu:2020}. Moreover, this test is relatively time-consuming, which is not desirable as the positive COVID-19 cases should be identified and tracked as fast as possible~\cite{Wang:2020}. Images~\cite{Parnian:SPM} in COVID-19 patients, on the other hand, have shown specific findings, such as ground-glass opacities with rounded morphology and a peripheral  lung distribution. Although imaging studies and theirs results can be obtained in a timely fashion, the previously described imaging finding may be seen in other viral or fungal infections or other entities such as organizing pneumonia, which limits the specificity of images and reduces the accuracy of a human-centered diagnosis.

\noindent
\textbf{Literature Review:}
Since revealing the potentials of computed tomography (CT) scans and X-ray images in detecting COVID-19 and weakness of the human-centered diagnosis, there have been several studies~\cite{Gozes:2020}\nocite{Narin:2020}-\cite{Farooq:2020} trying to develop automatic COVID-19 classification systems, mainly using Convolutional Neural Networks (CNNs)~\cite{Yamashita:2018}. Xu \textit{et al.}~\cite{Xu:2020} have first adopted a pre-trained 3D CNN to extract potential infected regions from the CT scans. These candidates are subsequently fed to a second CNN to classify them into three groups of COVID-19, Influenza-A-viral-pneumonia, and irrelevant-to-infection, with an overall accuracy of $86.7\%$. Wang \textit{et al.}~\cite{Wang:2020} have first extracted candidates using a threshold-based strategy. Consequently, for each case two or three regions are randomly selected to form the dataset. A pre-trained CNN is fine-tuned using the developed dataset. Finally, features are extracted from the CNN and fed to an ensemble of classifiers for the COVID-19 prediction, reaching an accuracy of $88\%$. CT scans are also utilized in Reference~\cite{Li:2020} to identify positive COVID-19 cases, where all slices are separately fed to the model and outputs are aggregated using a Max-pooling operation, reaching a sensitivity of~$90\%$. In a study by Wang and Wong~\cite{Wong:2020}, a CNN model is first pre-trained on the ImageNet dataset~\cite{Alex:2012}, followed by fine-tuning using a dataset of X-ray images to classify subjects as normal, bacterial, non-COVID-19 viral, and COVID-19 viral infection, achieving an overall accuracy of $83.5\%$. In a similar study by Sethy and Behera~\cite{Sethy:2020}, different CNN models are trained on X-ray images, followed by a Support Vector Machine (SVM) classifier to identify positive COVID-19 cases, reaching an accuracy of $95.38\%$.

\noindent
\textbf{Contributions:}
All the studies on deep learning-based COVID-19 classification have so far utilized CNNs, which although being powerful image processing techniques, are prone to an important drawback. They are unable to capture spacial relations between image instances. As a result of this inability, CNNs  cannot recognize the same object when it is rotated or subject to another type of transformation. Adopting a big dataset, including all the possible transformations, is the solution to this problem. However, in medical imaging problems, including the COVID-19 classification, huge datasets are not easily accessible. In particular, COVID-19 has been identified only recently, and large enough datasets are not yet developed. Capsule Networks (CapsNets)~\cite{Hinton:2018} are alternative models that are capable of capturing spatial information using routing by agreement, through which Capsules try to reach a mutual agreement on the existence of the objects. This agreement leverages the information coming from instances and object parts, and is therefore able to recognize their relations, without a huge dataset. Through several studies\cite{Afshar:2018}\nocite{Afshar2:2019,Afshar7:2019,Afshar1:2020,Afshar2:2020}-\cite{Afshar3:2019}, we have shown the superiority of the CapsNets for different medical problems such as brain tumor~\cite{Afshar:2018}\nocite{Afshar2:2019,Afshar7:2019,Afshar1:2020}-\cite{Afshar2:2020} and lung tumor classification~\cite{Afshar3:2019}. In this study, we propose a Capsule Network-based framework, referred to as the COVID-CAPS, for COVID-19 identification using X-ray images.  The proposed COVID-CAPS achieved an accuracy of $95.7\%$, a sensitivity of $90\%$, specificity of $95.8\%$, and Area Under the Curve (AUC) of $0.97$.

To potentially and further improve diagnosis capabilities of the COVID-CAPS, we considered pre-training and transfer learning using an external dataset of X-ray images, consisting of $94,323$ frontal view chest X-ray images for common thorax diseases. This dataset is extracted from the NIH Chest X-ray dataset~\cite{Wang:2017} including $112,120$ X-ray images for $14$ thorax abnormalities. From existing $15$ diseases in this dataset, $5$ classes were constructed with the help of a thoracic radiologist, with 18 years of experience in thoracic imaging (A. O.). It is worth mentioning that our pre-training strategy is in contrary to that of Reference~\cite{Wong:2020} where pre-training is performed  based on natural images (ImageNet dataset). Intuitively speaking, pre-training based on an X-ray dataset of similar nature is expected to result in better transfer learning in comparison to the case where natural images were used for this purpose. In summary, pre-training with an external dataset of X-ray images further improved accuracy of COVID-CAPS to $98.3\%$, specificity to $98.6\%$, and AUC to $0.97$, however, with a lower sensitivity of $80\%$. Trained COVID-CAPS model is available publicly for open access at \href{https://github.com/ShahinSHH/COVID-CAPS}{https://github.com/ShahinSHH/COVID-CAPS}. To the best of our knowledge, this is the first study investigating applicability of the CapsNet for the problem at hand.

The rest of the manuscript is organized as follows: Section~\ref{sec:caps} briefly introduces the Capsule networks. The COVID-CAPS is presented in Section~\ref{sec:framework}. Utilized dataset for evaluation of the proposed COVID-CAPS, and our results are presented in Section~\ref{sec:EXP}. Finally, Section~\ref{sec:con} concludes the work.

\section{Capsule Networks}\label{sec:caps}
\begin{figure*}
\centering
\includegraphics[width=1\textwidth]{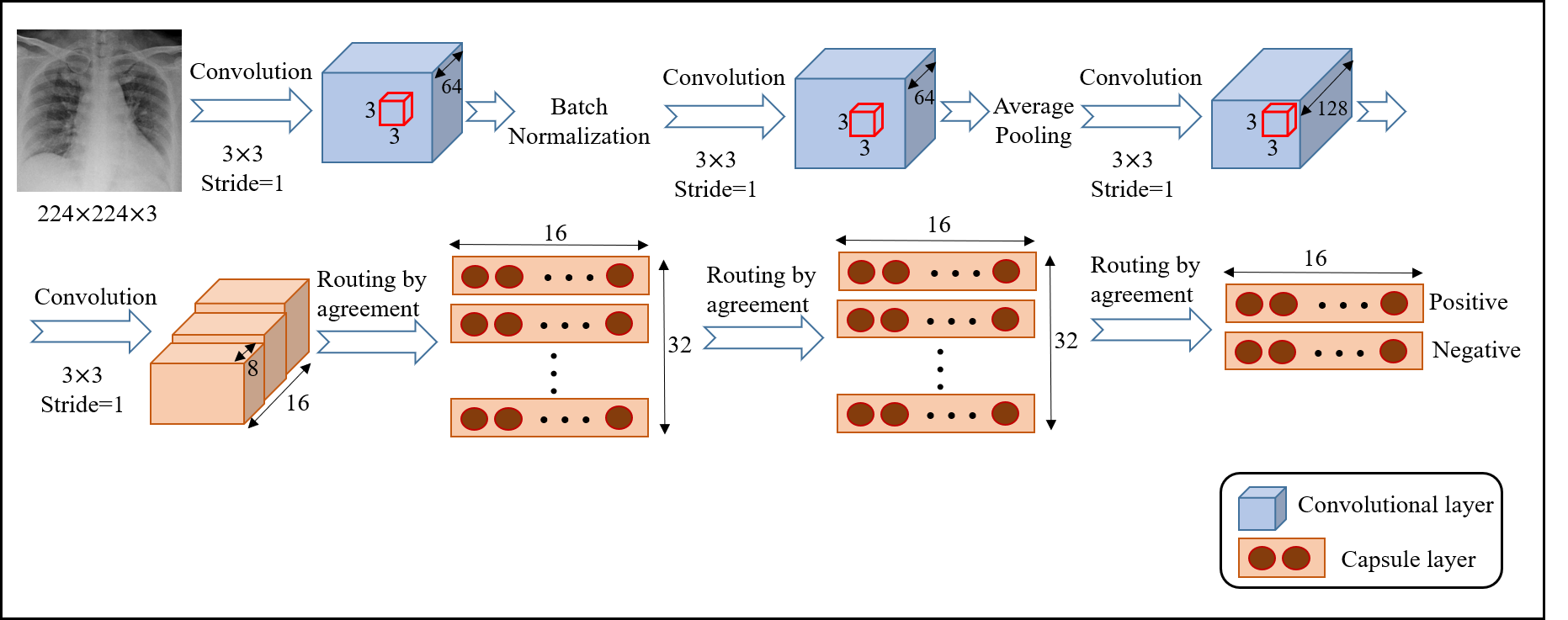}
\caption{ The proposed COVID-CAPS architecture. \label{fig:model}}
\vspace{-.1in}
\end{figure*}

Each layer of a Capsule Network (CapsNet) consists of several Capsules, each of which represents a specific image instance at a specific location, through several neurons. The length of a Capsule determines the existence probability of the associated instance. Similar to a regular CNN, each Capsule $i$, having the instantiation parameter $\u_i$, tries to predict the outputs of the next layer's Capsules, using a trainable weight matrix $\W_{ij}$, as follows
\begin{equation}
\hat{\u}_{j|i}=\W_{ij}\u_i,
\end{equation}
where $\hat{\u}_{j|i}$ denotes the prediction of Capsule $i$ for Capsule $j$. The predictions, however, are taken into account based on a coefficient, through the ``Routing by Agreement'' process, to determine the actual output of the Capsule $j$, denoted by $\s_j$, as follows
\begin{equation}
 a_{ij}=\s_j.\hat{\u}_{j|i},\\
\end{equation}
\begin{equation}
b_{ij}=b_{ij}+a_{ij},\\
\end{equation}
\begin{equation}
c_{ij}=\frac{\exp(b_{ij})}{\sum_k \exp(b_{ik})},\\
\end{equation}
\begin{equation}
\text{ and  }\s_j=\sum_ic_{ij}\hat{\u}_{j|i},
\end{equation}
where $a_{ij}$ denotes the agreement between predictions and outputs, and $c_{ij}$ is the score given to the predictions. In other words, this score determines the contribution of the prediction to the output. Routing by agreement is what makes the CapsNet different from a CNN and helps it identify the spatial relations.

The CapsNet loss function, $l_k$, associated with Capsule $k$, is calculated as follows
\begin{equation}
l_k=T_k\max(0,m^+-||\s_k||)^2+\lambda(1-T_k)\max(0,||\s_k||-m^-)^2,
\end{equation}
where $T_k$ is one whenever the class $k$ is present and zero otherwise. Terms $m^+$, $m^-$, and $\lambda$ are the hyper parameters of the model. The final loss is the summation over all the $l_k$s. This completes a brief introduction to CapsuleNets, next we present the COVID-CAPS framework.

\section{The Proposed COVID-CAPS}\label{sec:framework}

The architecture of the proposed COVID-CAPS is shown in Fig.~\ref{fig:model}, which consists of $4$  convolutional layers and $3$ Capsule layers. The inputs to the network are 3D X-ray images. The first layer is a convolutional one, followed by batch-normalization. The second layer is also a convolutional one, followed by average pooling. Similarly, the third and forth layers are convolutional ones, where the forth layer is reshaped to form the first Capsule layer. Consequently, three Capsule layers are embedded in the COVID-CAPS to perform the routing by agreement process. The last Capsule layer contains the instantiation parameters of the two classes of positive and negative COVID-19. The length of these two Capsules represents the probability of each class being present.

Since we have developed a Capsule Network-based architecture, which does not need a large dataset, we did not perform any data augmentation. However, since the number of positive cases, $N^+$, are less than the negative ones, $N^-$, we modified the loss function to handle the class imbalance problem. In other words, more weight is given to positive samples in the loss function, where weights are determined based on the proportion of the positive and negative cases, as follows
\begin{equation}
\text{loss}= \frac{N^+}{N^++N^-}\times \text{loss}^- +\frac{N^-}{N^++N^-}\times \text{loss}^+,
\end{equation}
where $\text{loss}^+$ denotes the loss associated with positive samples, and $\text{loss}^-$ denotes the loss associated with negative samples.

As stated previously, to potentially and further improve diagnosis capabilities of the COVID-CAPS, we considered pre-training the model in an initial step. In contrary to Reference~\cite{Wong:2020} where ImageNet dataset~\cite{Alex:2012} is used for pre-training, however, we constructed and utilized an X-ray dataset. The reason for not using ImageNet for pre-training is that the nature of images (natural images) in that dataset is totally different from COVID-19 X-ray dataset. It is expected that using a model pre-trained on  X-ray images of similar nature would result in better boosting of the COVID-CAPS.
For pre-training with an external dataset, the whole COVID-CAPS model is first trained on the external data, where the number of final Capsules is set to the number of output classes in the external set. From existing $15$ disease in the external dataset, $5$ classes were constructed with the help of a thoracic radiologist, with 18 years of experience in thoracic imaging (A. O.). To fine-tune the model using the COVID-19 dataset, the last Capsule layer is replaced with two Capsules to represent positive and negative COVID-19 cases. All the other Capsule layers are fine-tuned, whereas the conventional layers are fixed to the weights obtained in pre-training.

We used Adam optimizer with an initial learning rate of $10^{-3}$, $100$ epochs, and a batch size of $16$. We have split the training dataset, described in Section~\ref{sec:EXP}, into two sets of training ($90\%$) and validation ($10\%$), where training set is used to train the model and the validation set is used to select a model that has the best performance. Selected model is then tested on the testing set, for the final evaluation. The following four metrics are utilized to represent the performance: Accuracy; Sensitivity; Specificity, and Area Under the Curve (AUC). Next, we present the obtained results.
\section{Experimental Results} \label{sec:EXP}
\begin{table*}[t!]
\centering
\caption{ Results obtained from the proposed COVID-CAPS, along with the results from Reference~\cite{Sethy:2020}.}
\label{tab:res}
\vspace{.1in}
\begin{tabular}{|c|c|c|c|c|}
\hline
\multirow{2}{*}{\textbf{Method}} & \multirow{2}{*}{\textbf{Accuracy}} & \multirow{2}{*}{\textbf{Sensitivity}} & \multirow{2}{*}{\textbf{Specificity}} & \multirow{2}{*}{\textbf{Number of Trainable Parameters}} \\
&&&&\\
\hline

\multirow{2}{*}{\textbf{COVID-CAPS without pre-training}}   & \multirow{2}{*}{95.7\%}  & \multirow{2}{*}{90\%} &  \multirow{2}{*}{95.8\%} & \multirow{2}{*}{\textbf{295,488}} \\
&&&&\\
\hline
\multirow{2}{*}{\textbf{Pre-trained COVID-CAPS}}    & \multirow{2}{*}{\textbf{98.3\%}}  &\multirow{2}{*}{80\%} & \multirow{2}{*}{\textbf{98.6\%}} &  \multirow{2}{*}{\textbf{295,488}}  \\
&&&&\\
\hline
\multirow{2}{*}{\textbf{Reference~\cite{Sethy:2020}}}    & \multirow{2}{*}{95.38\%}  &\multirow{2}{*}{\textbf{97.29\%}} & \multirow{2}{*}{93.47\%} &  \multirow{2}{*}{23,000,000} \\
&&&&\\
\hline
\end{tabular}
\end{table*}

\begin{table*}[t!]
\centering
\caption{Description of the External X-ray images dataset used for pre-training COVID-CAPS.}
\label{tab:ex}
\vspace{.1in}
\begin{tabular}{|c|c|c|}
\hline
\textbf{Final Category} & \textbf{Initial Categories} & \textbf{Number of Images} \\
\hline
No Findings	  & No Findings	 & $60361$ \\
\hline
Tumors	 & Infiltration, Mass, Nodule & $16103$\\
\hline
Pleural Diseases	 & Effusion, Pleural Thickening, Pneumothorax & $8042$\\
\hline
Lung Infection	 & Consolidation, Pneumonia & $1668$\\
\hline
Others	 & Atelectasis, Cardiomegaly, Edema, Emphysema, Fibrosis, Hernia & $8149$\\
\hline
\end{tabular}
\end{table*}

\begin{figure}[t!]
\centering
\includegraphics[width=0.4\textwidth]{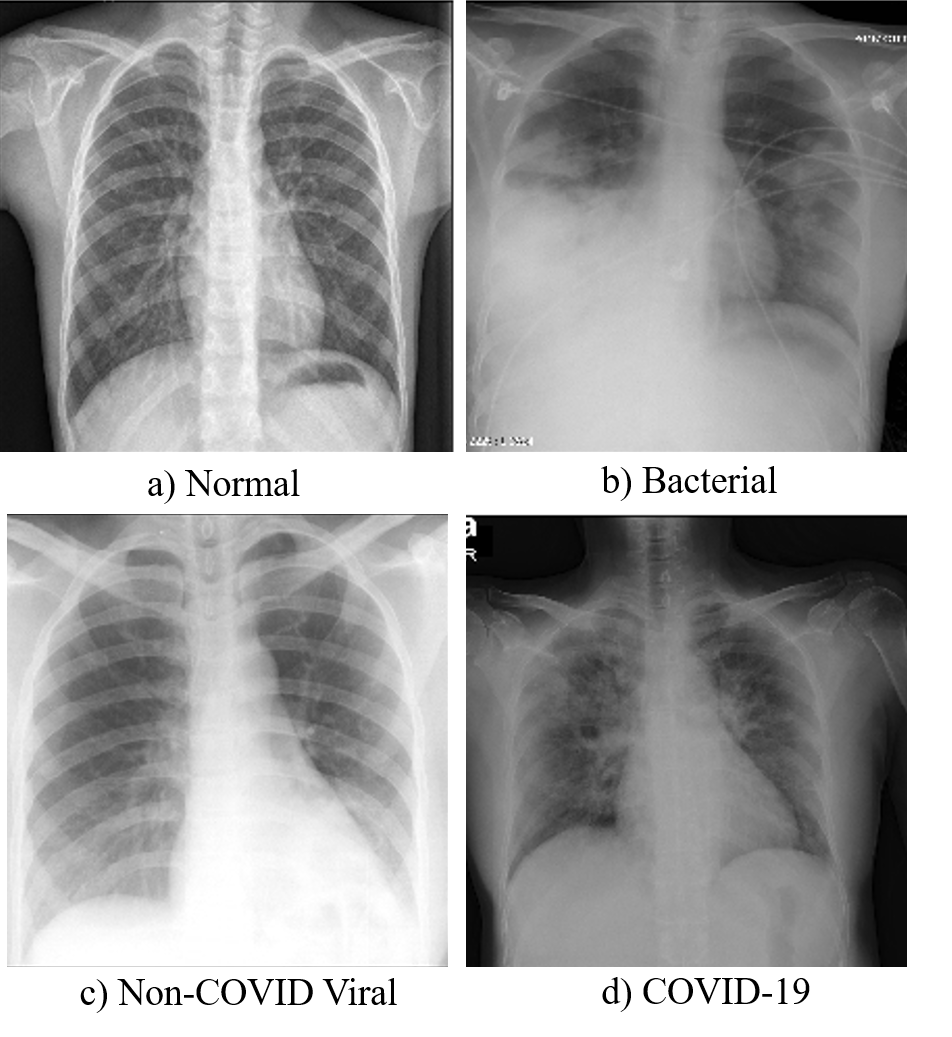}
\caption{Labels available in the dataset. \label{fig:data}}
\end{figure}

\begin{figure}[t!]
\centering
\includegraphics[width=0.4\textwidth]{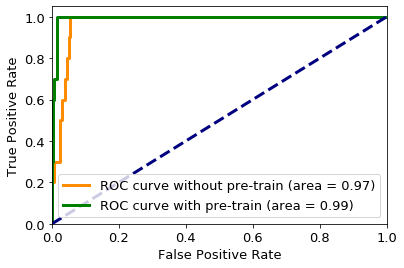}
\caption{ROC curve from the proposed COVID-CAPS. \label{fig:roc}}
\end{figure}

To conduct our experiments, we used the same dataset as Reference~\cite{Wong:2020}. This dataset is generated from two publicly available chest X-ray datasets~\cite{Cohen:2020,Mooney:2020}. As shown in Fig.~\ref{fig:data}, the generated dataset contains four different labels, i.e., Normal; Bacterial; Non-COVID Viral, and; COVID-19. As the main goal of this study is to identify positive COVID-19 cases, we binarized the labels as either positive or negative. In other words, the three labels of normal, bacterial, and non-COVID viral together form the negative class.

Using the aforementioned dataset, the proposed COVID-CAPS achieved an accuracy of $95.7\%$, a sensitivity of $90\%$, specificity of $95.8\%$, and AUC of $0.97$. The obtained receiver operating characteristic (ROC) curve is shown in Fig.~\ref{fig:roc}. \textit{In particular,  false positive cases have been further investigated to have an insight on what types are more subject to being mis-classified by COVID-19. It is observed that $54\%$ of the false positives are normal cases, whereas bacterial and non-COVID cases form only $27\%$ and $19\%$ of the false positives, respectively.}

As shown in Table~\ref{tab:res}, we compare our results with Reference~\cite{Sethy:2020} that has used the binarized version of the same dataset. COVID-CAPS outperforms its counterpart in terms of accuracy and specificity. Sensitivity is higher in the model proposed in Reference~\cite{Sethy:2020}, that contains $23$ million trainable parameters. Reference\cite{Narin:2020} is another study on the binarized version of the same X-ray images. However, as the negative label contains only normal cases (in contrast to including all normal, bacterial, and non-covid viral cases as negative), we did not compare the performance of the COVID-CAPS with this study. \textit{It is worth mentioning that the proposed COVID-CAPS has only $295,488$ trainable parameters. Compared to $23$ million trainable parameters of  the model proposed in Reference~\cite{Sethy:2020}, therefore,  COVID-CAPS can be trained and used in a more timely fashion, and eliminates the need for availability of powerful computational resources.}

In another experiment, we pre-trained the proposed COVID-CAPS using an external dataset of X-ray images, consisting of $94,323$ frontal view chest X-ray images for common thorax diseases. This dataset is extracted from the NIH Chest X-ray dataset~\cite{Wang:2017} including $112,120$ X-ray images for $14$ thorax abnormalities. This dataset also contains normal cases without specific findings in their corresponding images. In order to reduce the number of categories, we classified these $15$ groups into 5 categories based on the relations between the abnormalities in each disease. The first four groups are dedicated to No findings, Tumors, Pleural diseases, and Lung infections categories. The fifth group encompasses other images without specific relations with the first four groups. We then removed $17,797$ cases with multiple labels (appeared in more than one category) to reduce the complexity. The adopted dataset is then used to pre-train our model. Table~\ref{tab:ex} demonstrates our classification scheme and distribution of the data. Results obtained from fine-tuning the pre-trained COVID-CAPS is also shown in Table~\ref{tab:res}, according to which, pre-training improves accuracy and specificity.  The ROC curve is shown in Fig.~\ref{fig:roc}, according to which, the obtained AUC of $0.99$ outperforms that of COVID-CAPS without pre-training.

\section{Conclusion}  \label{sec:con}

In this study, we proposed a Capsule Network-based framework, referred to as the  COVID-CAPS, for diagnosis of COVID-19 from X-ray images. The proposed framework consists of several Capsule and convolutional layers, and the lost function is modified to account for the class-imbalance problem. The obtained results show that the COVID-CAPS has a satisfying performance with a low number of trainable parameters. Pre-training was able to further improve the accuracy, specificity, and AUC. Trained COVID-CAPS model is available publicly for open access at \href{https://github.com/ShahinSHH/COVID-CAPS}{https://github.com/ShahinSHH/COVID-CAPS}. As more and more COVID-19 cases are being identified all around the world, larger datasets are being generated. We will continue to further modify the architecture of the COVID-CAPS and incorporate new available datasets. New versions of the COVID-CAPS will be released upon development through the aforementioned link.

\section*{Acknowledgments}
This work was partially supported by the Natural Sciences and Engineering Research Council (NSERC) of Canada through the NSERC Discovery Grant RGPIN-2016-04988.



\begin{thebibliography}{10}

\bibitem{Xu:2020}
X. Xu, X. Jiang, {\em et al.},
\newblock ``Deep Learning System to Screen Coronavirus Disease 2019 Pneumonia,''
\newblock {\em arXiv:2002.09334}, 2020.


\bibitem{Wang:2020}
Sh. Wang, B. Kang, {\em et al.},
\newblock ``A Deep Learning Algorithm using CT Images to Screen for Corona Virus Disease (COVID-19),''
\newblock {\em medRxiv}, 2020.

\bibitem{Parnian:SPM}
P. Afshar, A. Mohammadi, K. N. Plataniotis, A. Oikonomou and H. Benali,
\newblock ``From Handcrafted to Deep-Learning-Based Cancer Radiomics: Challenges and Opportunities,"
\newblock {\em IEEE Signal Processing Magazine}, vol. 36, no. 4, pp. 132-160, July 2019.

\bibitem{Yamashita:2018}
R. Yamashita, M. Nishio, {\em et al.},
\newblock ``Convolutional Neural Networks: An Overview and Application in Radiology,''
\newblock {\em Insights into Imaging}, vol. 9, no. 4, pp. 611-629, 2018.


\bibitem{Gozes:2020}
O. Gozes, M.  Frid-Adar, {\em et al.},
\newblock ``Rapid AI Development Cycle for the Coronavirus (COVID-19) Pandemic:  Initial Results for Automated Detection \& Patient Monitoring  using Deep Learning CT Image Analysis ,''
\newblock {\em arXiv:2003.05037}, 2020


\bibitem{Narin:2020}
A. Narin, C. Kaya, Z. Pamuk ,
\newblock ``Automatic Detection of Coronavirus Disease (COVID-19) Using X-ray Images and Deep Convolutional Neural Networks ,''
\newblock {\em 	arXiv:2003.10849}, 2020

\bibitem{Farooq:2020}
M. Farooq, A. Hafeez,
\newblock ``COVID-ResNet: A Deep Learning Framework for Screening of COVID19 from Radiograph,''
\newblock {\em arXiv:2003.14395}, 2020

\bibitem{Li:2020}
L. Li, L. Qin, {\em et al.},
\newblock ``Artificial Intelligence Distinguishes COVID-19 from Community Acquired Pneumonia on Chest CT,''
\newblock {\em Radiology}, 2020

\bibitem{Wong:2020}
L. Wang, A. Wong,
\newblock ``COVID-Net: A Tailored Deep Convolutional Neural Network Design for Detection of COVID-19 Cases from Chest Radiography Images,''
\newblock {\em arXiv:2003.09871}, 2020.

\bibitem{Alex:2012}
A. Krizhevsky, I. Sutskever, G.~E. Hinton,
\newblock 	``ImageNet Classification with Deep Convolutional Neural Networks,''
\newblock {\em Neural Information Processing Systems (NIPS)} 2012.


\bibitem{Sethy:2020}
P.~K. Sethy, S.~K. Behera,
\newblock ``Detection of Coronavirus Disease (COVID-19) Based on Deep Features,''
\newblock {\em Preprints 2020, 2020030300}, 2020.

\bibitem{Hinton:2018}
G. Hinton, S. Sabour, N. Frosst,
\newblock `` Matrix Capsules With EM Routing,''
\newblock {\em ICLR}, 2018.
.
\bibitem{Afshar:2018}
P. Afshar, A.~Mohammadi, K.~N.  Plataniotis,
\newblock ``Brain Tumor Type Classification via Capsule Networks,''
\newblock {\em IEEE International Conference on Image Processing (ICIP)}, pp. 3129-3133, 2018.


  \bibitem{Afshar2:2019}
P. Afshar, K.~N Plataniotis, A.~Mohammadi,
\newblock ``Capsule Networks for Brain Tumor Classification Based on Mri Images and Coarse Tumor Boundaries,''
\newblock {\em ICASSP 2019-2019 IEEE International Conference on Acoustics, Speech and Signal Processing (ICASSP)}, pp. 1368-1372, 2019.

  \bibitem{Afshar7:2019}
P. Afshar, K.~N Plataniotis, A.~Mohammadi,
\newblock ``Capsule Networks' Interpretability for Brain Tumor Classification Via Radiomics Analyses,''
\newblock {\em 2019 IEEE International Conference on Image Processing (ICIP)}, pp. 13816-3820, 2019.

  \bibitem{Afshar1:2020}
P. Afshar, K.~N Plataniotis, A. Mohammadi
\newblock ``BoostCaps: A Boosted Capsule Network for Brain Tumor Classification,''
\newblock Accepted in {\em IEEE Engineering in Medicine and Biology Society (EMBC)}, 2020.

  \bibitem{Afshar2:2020}
P. Afshar, K.~N Plataniotis, A. Mohammadi
\newblock ``A Bayesian Approach to Brain Tumor Classification Using Capsule Networks,''
\newblock Submitted to {\em IEEE International Conference on Image Processing (ICIP)}, 2020.


\bibitem{Afshar3:2019}
P. Afshar, A. Oikonomou, P.N. Tyrrell, K. Farahani, K.~N Plataniotis, A.~Mohammadi,
\newblock ``3D-MCN: A 3D Multi-Scale Capsule Network for Lung Nodule Malignancy Classification,''
\newblock  Accepted with minor revision in {\em Scientific Reports}.


  \bibitem{Cohen:2020}
J.~P.  Cohen,
\newblock ``Covid Chest x-ray Dataset,''
\newblock {\em https://github.com/ieee8023/covid-chestxray-dataset}, 2020.

  \bibitem{Mooney:2020}
P.  Mooney,
\newblock ``Kaggle Chest x-ray Images (Pneumonia) Dataset,''
\newblock {\em https://github.com/ieee8023/covid-chestxray-dataset}, 2020.

  \bibitem{Wang:2017}
X. Wang, Y. Peng, {\em et al.},
\newblock ``ChestX-ray8: Hospital-scale Chest X-ray Database and Benchmarks on Weakly-Supervised Classification and Localization of Common Thorax Diseases,''
\newblock {\em 2017 IEEE Conference on Computer Vision and Pattern Recognition (CVPR)}, pp. 3462-3471, 2017.

\end{thebibliography}
\end{document}